\documentclass[3p,times,procedia]{elsarticle}
\usepackage{graphicx}
\usepackage{amsmath}
\usepackage{xcolor}
\usepackage{subcaption}

\usepackage[bookmarks=false]{hyperref}
    \hypersetup{colorlinks,
      linkcolor=blue,
      citecolor=blue,
      urlcolor=blue}

\usepackage{amssymb}

\usepackage[figuresright]{rotating}

\begin{document}
\begin{frontmatter}





\title{Fuzzy Approach for Audio-Video Emotion Recognition in Computer Games for Children}


\author[a]{Pavel Kozlov} 
\author[a]{Alisher Akram} 

\author[a]{Pakizar Shamoi\corref{cor1}}

\address[a]{School of Information Technology and Engineering, Kazakh-British Technical University, Almaty, Kazakhstan}

\begin{abstract}
Computer games are widespread nowadays and enjoyed by people of all ages. But when it comes to kids, playing these games can be more than just fun—it's a way for them to develop important skills and build emotional intelligence. Facial expressions and sounds that kids produce during gameplay reflect their feelings, thoughts, and moods. In this paper, we propose a novel framework that integrates a fuzzy approach for the recognition of emotions through the analysis of audio and video data. Our focus lies within the specific context of computer games tailored for children, aiming to enhance their overall user experience. We use the FER dataset to detect facial emotions in video frames recorded from the screen during the game. For the audio emotion recognition of sounds a kid produces during the game, we use CREMA-D, TESS, RAVDESS, and Savee datasets. Next, a fuzzy inference system is used for the fusion of results. Besides this, our system can detect emotion stability and emotion diversity during gameplay, which, together with prevailing emotion report, can serve as valuable information for parents worrying about the effect of certain games on their kids. The proposed approach has shown promising results in the preliminary experiments we conducted, involving 3 different video games, namely fighting, racing, and logic games, and providing emotion-tracking results for kids in each game. Our study can contribute to the advancement of child-oriented game development, which is not only engaging but also accounts for children's cognitive and emotional states.

\end{abstract}

\cortext[cor1]{Corresponding author. Tel.: +7-701-349-0001.}
\end{frontmatter}





\section{Introduction}



Many parents are concerned about how much time their children spend playing computer games, and concerns regarding the effects of video games on aspects like mental health and cognitive abilities have become consistent topics in societal dialogues \cite{psyc}. A majority of parents, specifically 64\%, held the belief that video games were responsible for fostering addiction. Furthermore, more than one out of every five parents were worried about video games affecting their own child \cite{Frontier}.

At the same time, most parents recognize the benefits of games and allow children to download applications \cite{rideot}. It has long been assumed that emotions have a strong influence on human behavior, actions and mental abilities \cite{Metcalfe1999}. Many teachers are convinced that the correct handling of games can be useful for the development of a child as well. The development of cognitive skills based on games depends on the quality of the content offered by the developers of such games. The emotional state of the child during the game  largely determines the interest in the computer game.

\begin{figure}[t]
    \begin{minipage}{0.56\linewidth}
        \centering
        \includegraphics[height=1.9in]{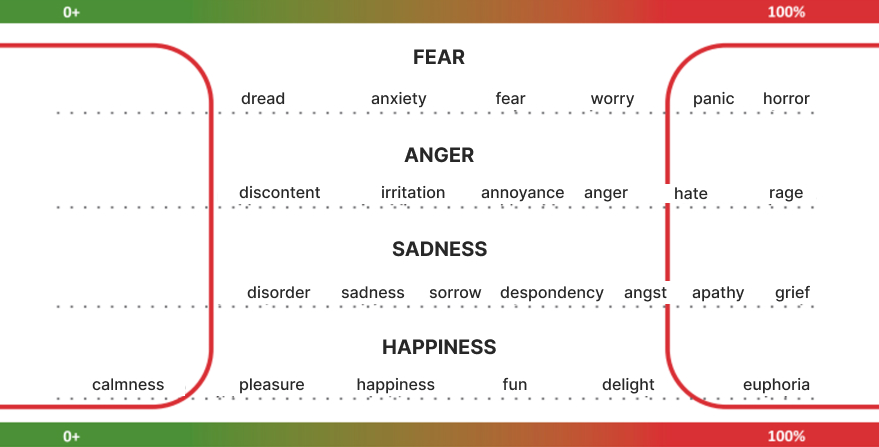}
        \captionof{figure}{Basic human emotions.}
        \label{fig:Figure2}
    \end{minipage}
    \hfill
    \begin{minipage}{0.4\linewidth}
        \centering
        \includegraphics[height=2.1in]{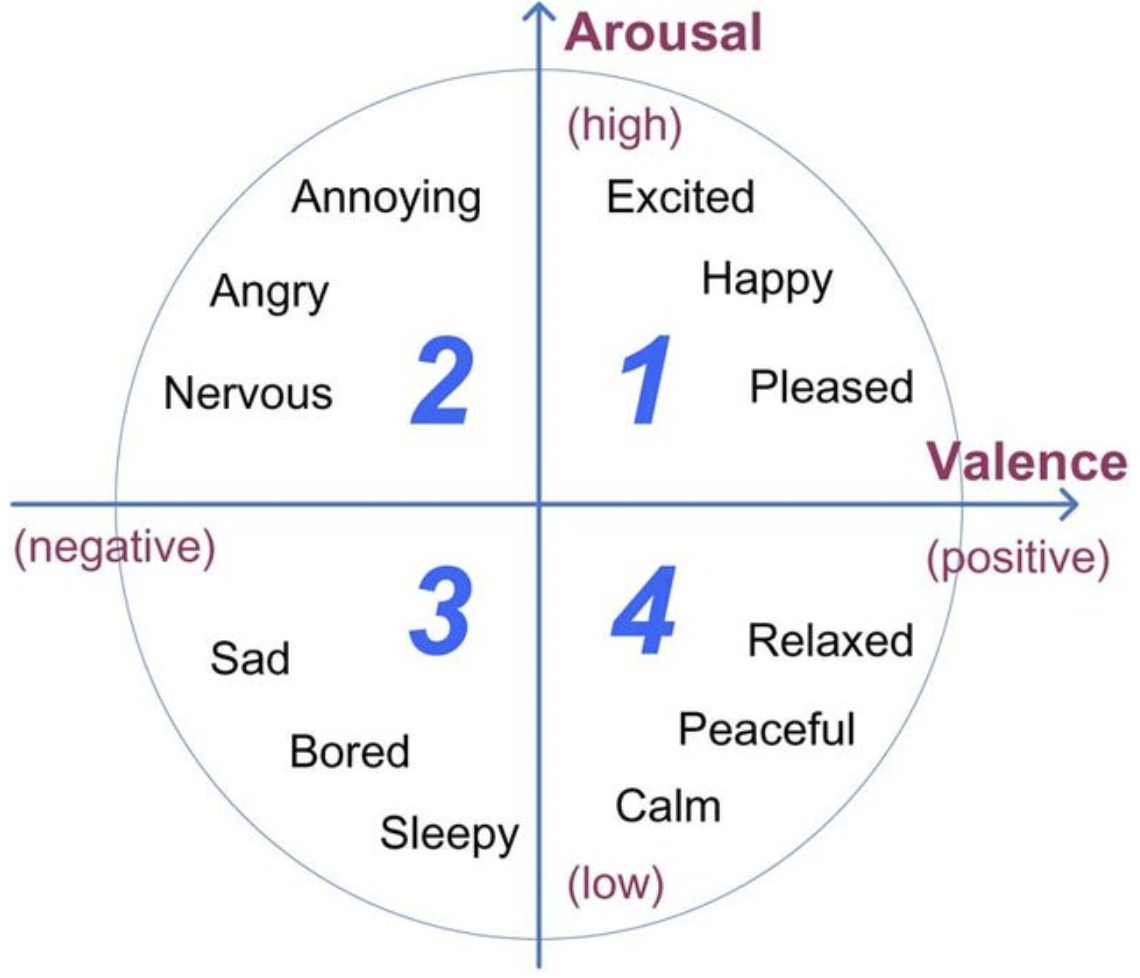}
        \captionof{figure}{Thayer's arousal-valence emotion plane \cite{thayer}, \cite{figure}}.
        \label{tab:Figure4}
    \end{minipage}
\end{figure}

There are six classic human emotions: happiness, surprise, fear, disgust, anger and sadness. However, according to recent findings \cite{CurrentBiology}, basic emotion transmission is divided into four (rather than six) types. Specifically, in the early phases, anger and disgust, as well as fear and surprise, are perceived identically. A wrinkled nose, for example, expresses both anger and disgust, while lifted eyebrows communicate surprise and fear. Basic human emotions are represented in Fig. \ref{fig:Figure2}. We use Thayer's arousal-valence emotion plane \cite{thayer} as our taxonomy and use seven emotions (six basic and neutral) belonging to one of the four quadrants of the emotion plane (See Fig. \ref{tab:Figure4}). In adults, the expression of emotions is less natural and determined by upbringing and cultural code. In general, the language of emotions is more universal, but there are some differences in facial expressions and gestures among different peoples.

The field of emotion recognition (ER) has gained growing interest in recent times. The complexity arising from factors like different poses, illumination conditions, motion blurring, and more makes the identification of emotions from audio-video sources a challenging task \cite{fusion2}. Moreover, limited number of works investigate kids' ER. One of the recent studies explored the use of ER to improve online school learning \cite{disc}.

Most emotion recognition algorithms are still limited to a single modality at the moment. However, in everyday life, humans frequently conceal their true feelings, which leads to the dilemma that single modal emotion recognition accuracy is relatively poor \cite{icanet}. Majority of works in this area employ CNN models for ER task. Some studies use multiple deep models (CNN, RNN for images and SVM, LSTM for acoustic features) \cite{fusion3}. Authors of the other study proposed multi-modal residual perceptron network for multimodal ER \cite{fusion1}. An interesting approach has been proposed for ER based on generating a representative set of frames from videos using the eigenspace domain and Principal component analysis \cite{articleeigen}. Another paper introduces Spatiotemporal attention-based multimodal deep neural networks for dimensional ER \cite{attention}.

Parents, wanting to occupy their child with something useful on a smartphone or computer, are not always sure about the effects of such games. Current cognitive learning games very often forget about the relationship between cognition and emotions, which is characterized by a "hot executive function". \cite{Gray2019}. Game developers are now more focused only on the end result, forgetting about the exciting gameplay and emotions that children experience in the process. The interaction of children with educational games and applications should be easily accessible and engaging, while encouraging them to achieve and complete tasks.

In this article, we introduce the framework that integrates a fuzzy logic approach to precisely identify emotions by analyzing both audio and video data. Our primary emphasis is on computer games designed for children, with the goal of improving their overall gaming experience. By automatically monitoring the emotions of children as they navigate through gameplay, developers can pinpoint pivotal moments within the gaming experience and can make games that truly connect with kids. Our contributions include proposed fuzzy fusion technique and exploring the audio-video emotional stability and diversity besides emotions.

\section{Methodology}
\begin{figure}[t]
\centerline{\includegraphics[width=\textwidth]{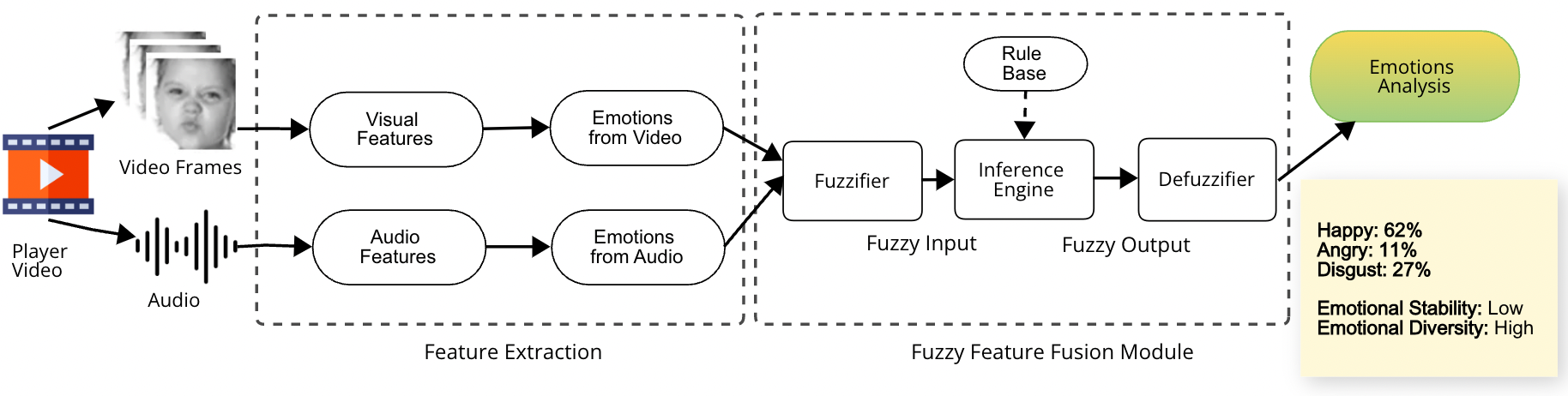}}
\caption{Methodology representation.}
\label{figmain}
\end{figure}

The proposed approach is shown in Fig. \ref{figmain}. Our framework has two important stages - feature extraction and emotion detection, fuzzy fusion of emotions. From incoming player video data we extract audio and video frames, perform feature extraction, detect emotions, and pass them to a fuzzy inference system to perform the fusion.

\subsection{FER 2013}
\begin{figure}[t]
    \begin{minipage}[b]{0.5\linewidth}
        \centering
        \includegraphics[width=3.1in]{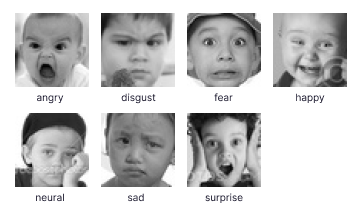}
        \captionof{figure}{Some samples from FER with children emotions}
        \label{fig:Figure1}
    \end{minipage}
    \hfill
    \begin{minipage}[b]{0.5\linewidth}
        \centering
        \includegraphics[width=3.1in]{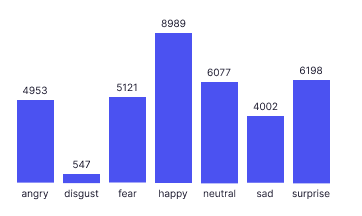}
        \captionof{figure}{Frequency  of emotions in a dataset}
        \label{tab:Figure3}
    \end{minipage}
\end{figure}

 We used facial expression recognition 2013 (FER 2013) emotion dataset,  \cite{ferr}, which was presented at the conference \cite{Kaggle}. This database contains 35887 black-and-white images of people's faces with a resolution of 48x48 pixels. All images were divided into 7 categories: \textit{0=Angry, 1=Disgust, 2=Fear, 3=Happy, 4=Sad, 5=Surprise, 6=Neutral}. An example of children's emotions from the database is shown in Fig. \ref{fig:Figure1}.

The dataset contains images not only of fully open faces but also partially closed, for example with a hand, low-contrast images, and images of people in glasses, etc. The dataset is divided into two parts, test and training. The test set is needed to compare the recognition accuracy among other models. The training set is needed for the training and optimization of models. The FER2013 dataset, divided into emotion categories, is shown in Fig. \ref{tab:Figure3}. The FER dataset is normalized in such a way that its data resembles a normal distribution: zero expectation and unit variance.  Output data is registered  in seven categories: "angry", "disgust", "fear", "happy", "sad", "surprise", "neutral". Each emotion is evaluated by the result on a scale from 0 to 1.

 The project was built on a convolutional neural network-based model \cite{GitHub}. You can also retrain the model if needed when calling and initializing the model. The accuracy of the CNN model on test data is 78\%. The constructor parameter is the Multitask Cascaded Neural Network (MTCNN) facial recognition technique. When the value is "True", the MTCNN model is used to detect faces, and when the value is set to "False", the function uses the OpenCV Haar Cascade classifier by default.

\subsection{Audio Emotion Recognition}
Audio, or Speech Emotion Recognition (SER) involves identifying human emotions and audio signals. Voice patterns frequently convey underlying emotions via variations in tone and pitch. In order to detect emotions in audio extracted from player video, we used Speech Emotion Recognition model \cite{audio_kaggle}, which was trained using the well-known datasets of audio clips annotated with emotions, namely, Crowd Sourced Emotional Multimodal Actors Dataset (CREMA-D, contains 7,442 audio clips) \cite{crema}, Toronto emotional speech set (TESS, 2800 audio files) \cite{tess}, Ryerson Audio-Visual Database of Emotional Speech and Song (RAVDESS, contains 1440 audio files) \cite{ravd}, Surrey Audio-Visual Expressed Emotion (Savee, 480 audio files) \cite{savee}. The accuracy of the model on test data is 60.74 \%.

\subsection{PyAutoGUI}
We use PyAutoGUI for automatic screening of the screen with the game and the child's face every second. PyAutoGUI is a cross-platform library for automating actions on a computer using Python scripts. With the help of this library, a screen with a logic game played by a child and a child's face will be displayed. 

\subsection{Fuzzy Sets and Logic}
There are some difficulties in recognizing emotions in faces, such as emotion ambiguity and a small number of emotion classes in comparison to human emotions \cite{aida}. Fuzzy logic is a powerful tool for handling  imprecision and it  was used in several studies to express emotions and their intensity \cite{peerj}. A fuzzy set is a class of objects that has a range of membership grades \cite{zadeh}. The main reason we used fuzzy sets and logic in our study is that they help us to rate emotions in a human-consistent manner because they do not have clearly defined bounds. Despite being less accurate, a language value is closer to human cognitive processes than a number \cite{fuzzycw}.

\begin{figure}[t]
\centerline{\includegraphics[width=6in]{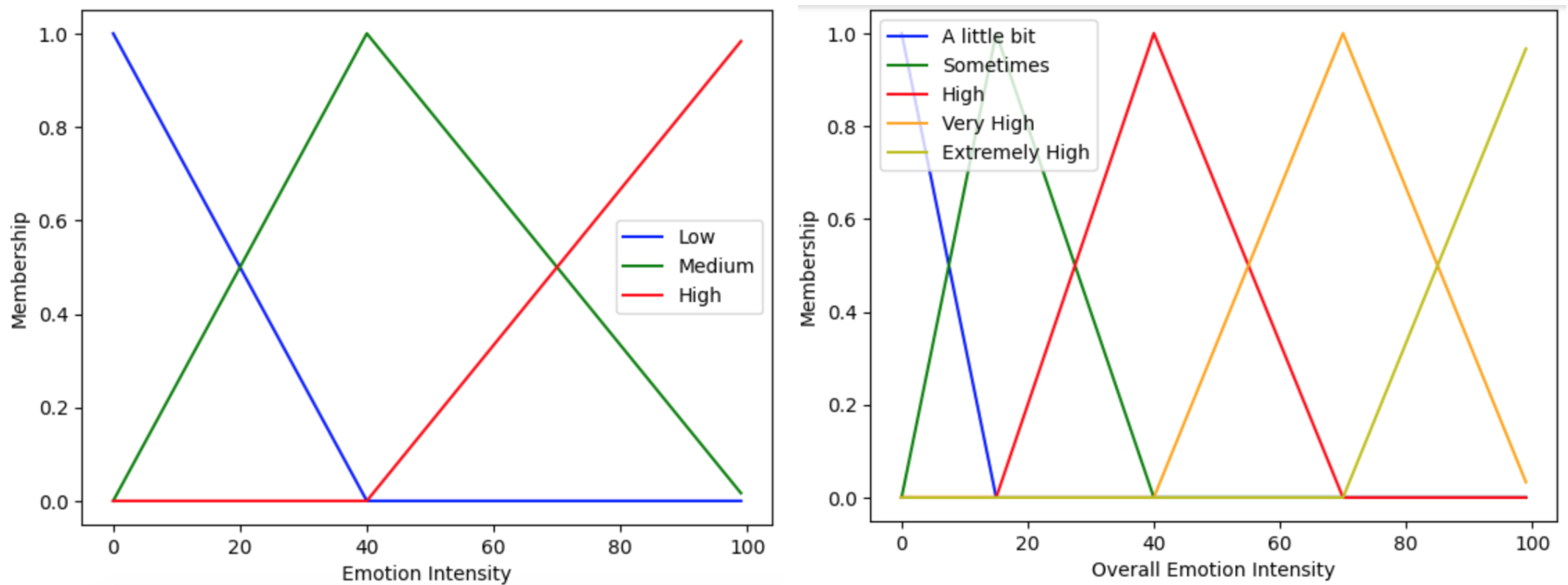}}
\caption{Input Fuzzy sets for \textit{Emotion Intensity} (the same for audio and video) and Output Fuzzy Sets for \textit{Overall Emotion Intensity}.}
\label{figsets}
\end{figure}

We partition the spectrum of possible emotions corresponding to linguistic tags \cite{jaciii}. We have two input variables for each emotion - Audio and Video Emotion Intensity.  The output variable is simply the \textit{Overall Emotion Intensity} in percentage points (see Fig. \ref{figsets}). As can be seen from the Fig, we have 'Low', "Medium, and 'High' fuzzy sets for both input variables and 'A little bit', 'Sometimes', 'High', 'Very High', and 'Extremely High' for the output variable.

 To build fuzzy relationships between input and output variables we use fuzzy rules. In our fuzzy inference system, we have nine fuzzy rules as shown in Table \ref{Table1}. The detailed example is provided in section \textit{Application and Results}.


\begin{table}[tb]
\caption{ Fuzzy rules used in the fuzzy inference system.}
\label{Table1}
\begin{tabular*}{\hsize}{@{\extracolsep{\fill}}llll@{}}
\textbf{Rules }& \textbf{Audio Emotion Intensity} & \textbf{Video Emotion Intensity} & \textbf{Overall Emotion Intensity} \\

1                           & Low                     & Low                     & Little Bit                \\
2                           & Low                     & Medium                  & Sometimes                 \\
3                           & Low                     & High                    & High                      \\
4                           & Medium                  & Low                     & Sometimes                 \\
5                           & Medium                  & Medium                  & High                      \\
6                           & Medium                  & High                    & Very High                 \\
7                           & High                    & Low                     & Sometimes                 \\
8                           & Medium                  & Medium                  & Very High                 \\
9                           & High                    & High                    & Extremely High   \\ 
\end{tabular*}
\end{table}

\section{Application and Results}

\subsection{Prototype Application}

\begin{figure}[tb]
\centerline{\includegraphics[height=2.6in]{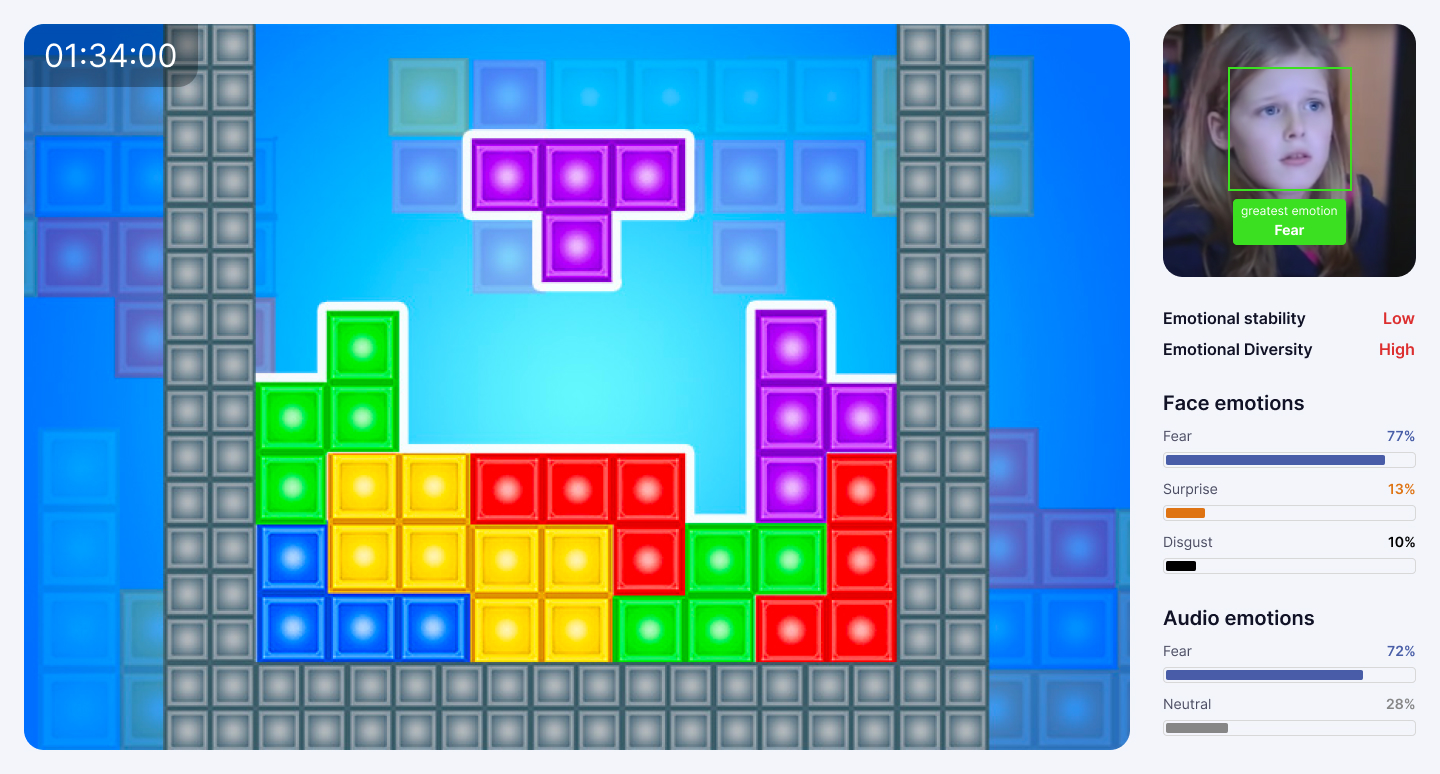}}
\caption{Example of the application interface.}
\label{figdesign}
\end{figure}

Fig. \ref{figdesign} illustrates the prototype application mockup. As it can be seen, it allows to track the average emotion, prevailing emotions in audio and video, emotional stability and diversity. As a result, parents will be offered a report about the feelings their child had while playing different computer games, emotional stability/instability and what were the prevailing emotions associated with this condition. This report would be really helpful for parents. It would let them see the effect of different games and how much their child can play. Together with a psychologist, it can help to figure out the best way to manage gaming based on their child's emotions.

\subsection{Experimental Results}
In this section we present our preliminary experimental results. We conducted an experiment on a 7-year-old child and looked at his emotions during three games: a \textit{fighting}, \textit{racing}, and \textit{logic} game. The speech emotion analysis led to the following results obtained on 10-sec audio extracted from the video and analyzed for each second:
\begin{itemize}
    \item Fight game - ['disgust' 'sad' 'sad' 'sad' 'disgust' 'happy' 'fearful' 'sad' 'sad' 'disgust']
    \item Racing game - ['sad' 'sad' 'sad' 'fearful' 'neutral' 'neutral' 'fearful' 'happy' 'sad' 'disgust']
    \item Logic Game - ['neutral' 'neutral' 'neutral' 'neutral' 'disgust' 'angry' 'neutral' 'neutral' 'sad' 'neutral']
\end{itemize}

\begin{figure}
  \begin{subfigure}{0.33\textwidth}
    \includegraphics[width=\linewidth]{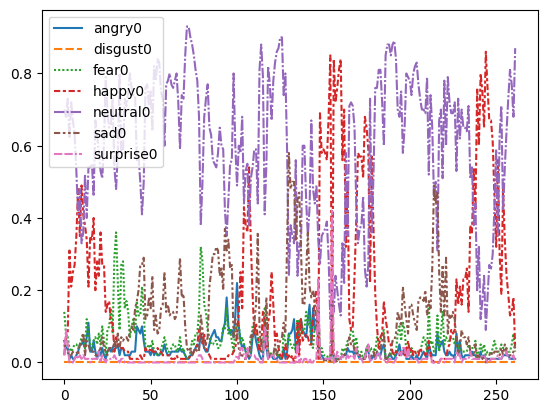}
    \caption{Fight Game} \label{fig:1a}
  \end{subfigure}%
  \hspace*{\fill}   
  \begin{subfigure}{0.33\textwidth}
    \includegraphics[width=\linewidth]{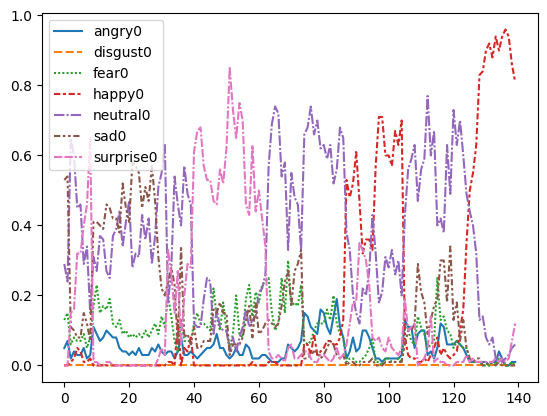}
    \caption{Racing Game} \label{fig:1b}
  \end{subfigure}%
  \hspace*{\fill}   
  \begin{subfigure}{0.33\textwidth}
    \includegraphics[width=\linewidth]{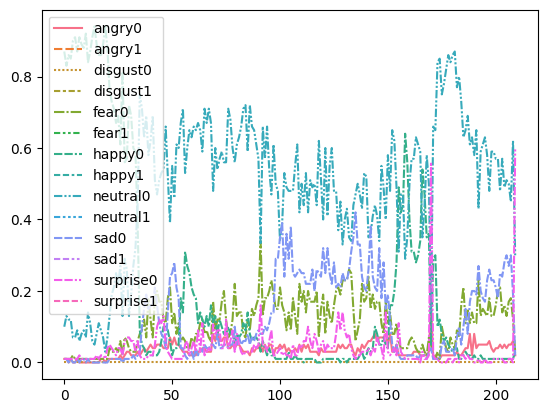}
    \caption{Logic Game (tic-tac-toe)} \label{fig:1c}
  \end{subfigure}
\caption{Video Emotion Recognition Results} \label{fig:1}
\label{abc}
\end{figure}

\begin{figure}[t]
\centerline{\includegraphics[width=16.5cm]{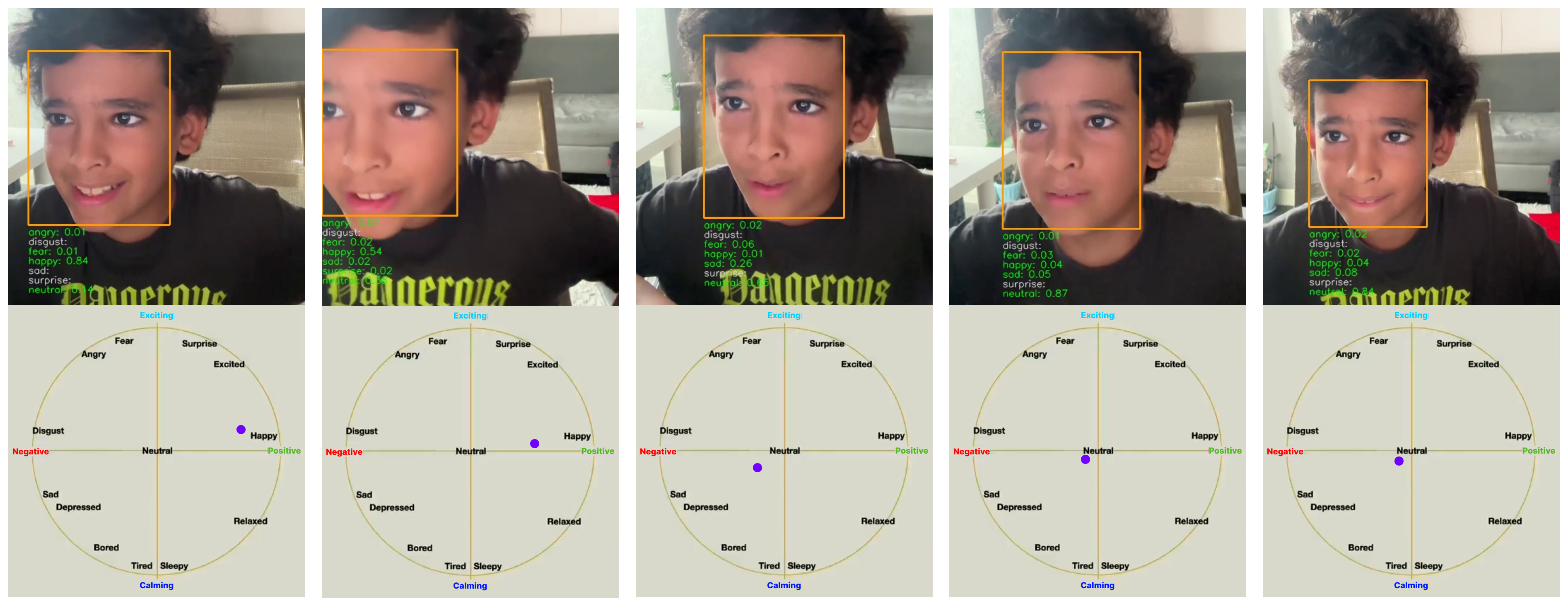}}
\caption{Video emotion detection results for the \textit{Fight} game and Thayer’s arousal-valence emotion planes for each of the 5 selected frames.}
\label{figsaliko}
\end{figure}

Video emotion recognition results for a \textit{Fight }game are shown in Fig. \ref{figsaliko}. Table \ref{frames} illustrates the emotion detection results for some video frames. To illustrate how emotions evolve, 5 exemplary frames (F1, F2, F3, F4, F5) were selected from the set of 262 frames. These frames match those in Fig. \ref{figsaliko}. Fig. \ref{abc} shows emotion tracking results for each of the selected games. As we can see from Fig. \ref{abc}, \textit{Fight} game data exhibits emotional instability and more emotional diversity than other games, including \textit{happy}, \textit{neutral}, \textit{sad}, and \textit{fear} emotions. We can also see that for a Logic game, there is one leading \textit{neutral} emotion. Emotional stability can be related with standard deviation (Table \ref{frames}) and emotional diversity - with number of emotions with \textit{High} or \textit{Medium} intensity.

\begin{table}[tb]
\caption{Emotions recognition results from video corresponding to a Fight game. 5 illustrative frames (F1, F2,F3, F4, F5) were chosen among 262 to show emotions change. These frames correspond to ones presented in Fig. \ref{figsaliko}.} 
\label{frames}
\begin{tabular*}{\hsize}{@{\extracolsep{\fill}}llllllllllllll@{}}
\textbf{Emotion}  & \textbf{F1} & \textbf{...} & \textbf{F2} & \textbf{...} & \textbf{F3} & \textbf{...} & \textbf{F4} & \textbf{...} & \textbf{F5} & \textbf{Mean} & \textbf{Median} & \textbf{Variance} & \textbf{SD} \\
\textbf{Happy}    & 0.04        & ...          & 0.01        & ...          & 0.84        & ...          & 0.54        & ...          & 0.04        & 0.18          & 0.07            & 0.05              & 0.23                   \\
\textbf{Angry}    & 0.01        & ... & 0.02        & ... & 0.01        & ... & 0.01        & ... & 0.02        & 0.04          & 0.03            & 0.001             & 0.03                   \\
\textbf{Disgust}  & 0           & ...          & 0           & ...          & 0           & ...          & 0           & ...          & 0           & 0.0           & 0.0             & 0.0               & 0.0                    \\
\textbf{Fear}     & 0.03        & ... & 0.06        & ... & 0.01        & ... & 0.02        & ... & 0.02        & 0.06          & 0.05            & 0.003             & 0.05                   \\
\textbf{Neutral}  & 0.87        & ...          & 0.65        & ...          & 0.14        & ...          & 0.38        & ...          & 0.84        & 0.59          & 0.63            & 0.04              & 0.2                    \\
\textbf{Sad}      & 0.05        & ... & 0.26        & ... & 0           & ... & 0.02        & ... & 0.08        & 0.11          & 0.07            & 0.01              & 0.11                   \\
\textbf{Surprise} & 0           & ...          & 0           & ...          & 0           & ...          & 0.02        & ...          & 0           & 0.01          & 0.01            & 0.001             & 0.03        \\       

\end{tabular*}
\end{table}


We can now simulate our fuzzy system by simply specifying the inputs and applying the defuzzification method. For example, let us find out what would be the overall emotion intensity in the following scenario: the audio and video intensity values for \textit{Happy} Emotion are 12\% and 85\% respectively. Then, the output membership functions are combined using the maximum operator (fuzzy aggregation). Next, to get a crisp answer, we need to perform defuzzification, for that we use a centroid method. As a result of performing aggregation based on fuzzy rules, we get 47.55 \% as the overall intensity for a Happy emotion. The visualized result is presented in Fig. \ref{fig:1}.

\begin{figure}

  \begin{subfigure}{0.32\textwidth}
    \includegraphics[width=\linewidth]{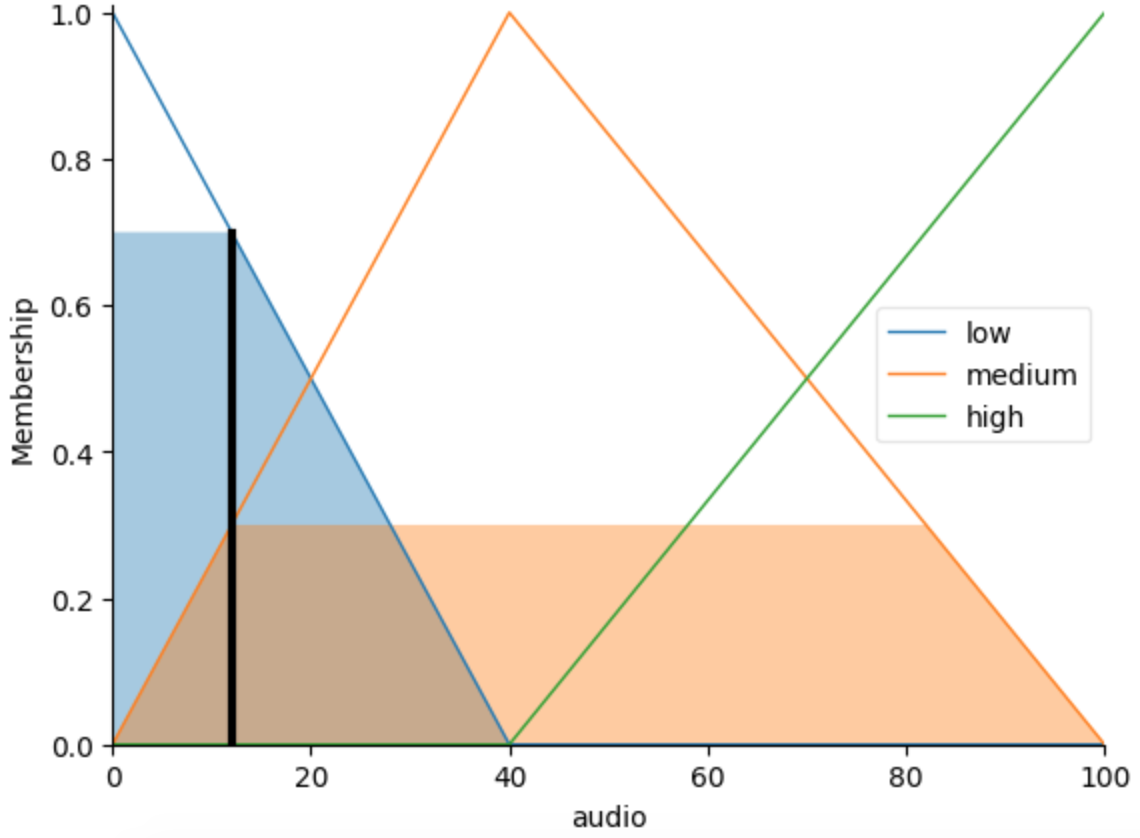}
   
    \caption{Applying input 12\% on \textit{Audio Intensity} fuzzy set} \label{fig:1a}
  \end{subfigure}%
  \hspace*{\fill}   
  \begin{subfigure}{0.32\textwidth}
    \includegraphics[width=\linewidth]{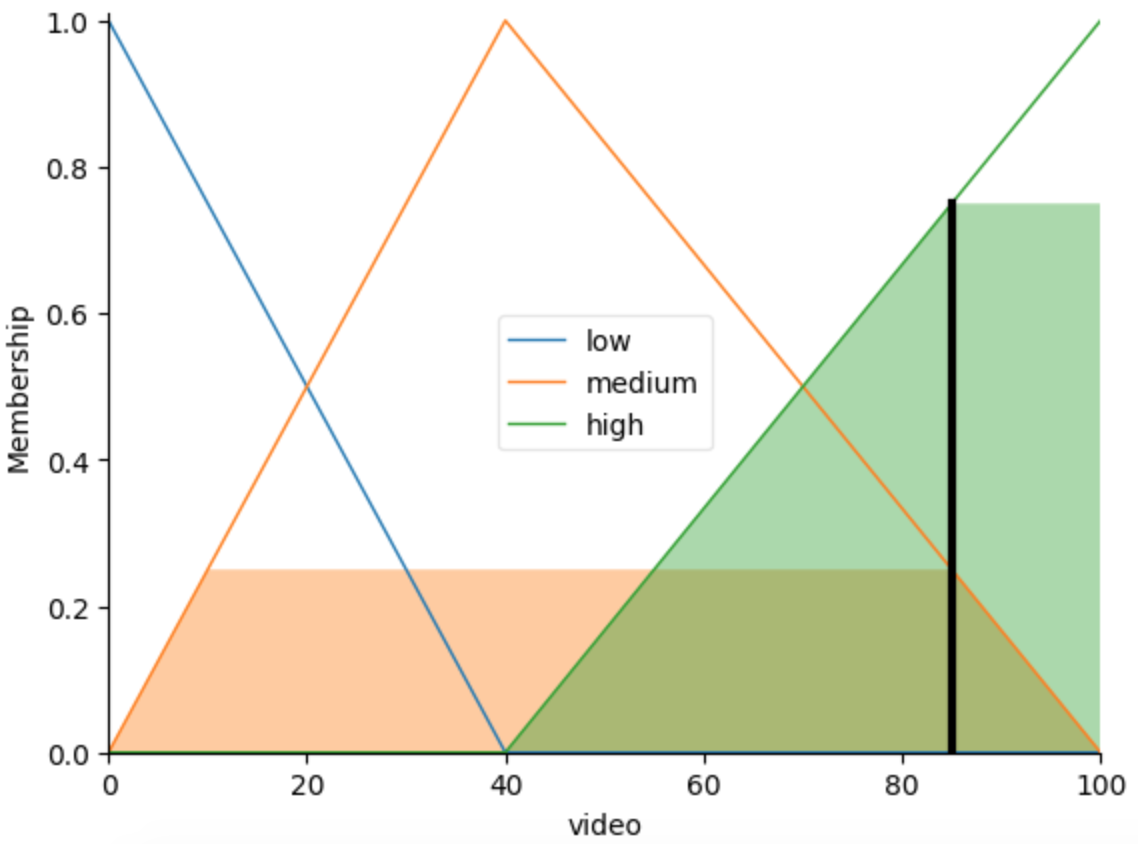}
    \caption{Applying input 85\% on \textit{Video Intensity} fuzzy set} \label{fig:1b}
  \end{subfigure}%
  \hspace*{\fill}   
  \begin{subfigure}{0.32\textwidth}
    \includegraphics[width=\linewidth]{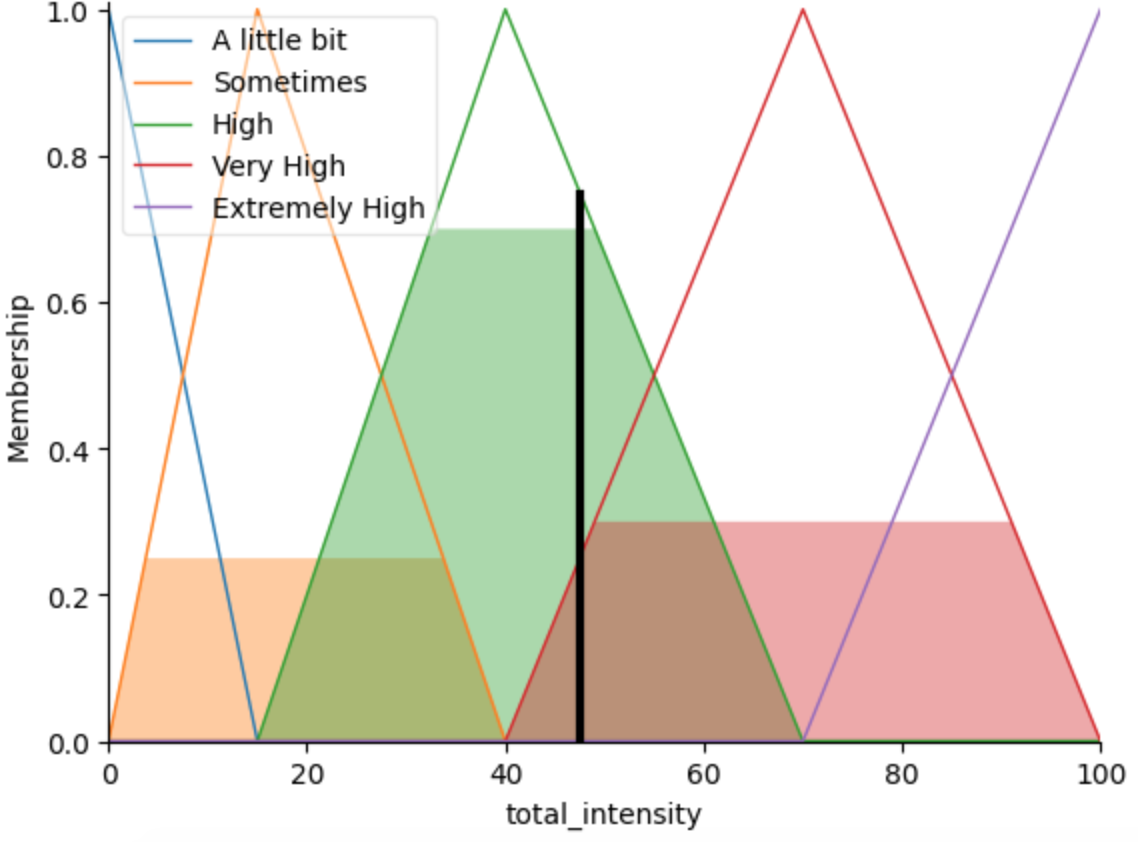}
    \caption{Aggregated Membership and Result, 47.55\%} 
  \end{subfigure}
\caption{Simulation Results.} \label{fig:1}

\end{figure}



\section{Conclusion}


Our study aimed to highlight the significance of audio-video ER in developing more engaging, safe, and effective games for children. We proposed a fuzzy logic-based approach to aggregate the emotions detected from video frames and sounds. For that, we used FER emotion library as the basis.

Our work can contribute to the advancement of emotionally aware computer games. Using ER software, game developers can identify problems and work on their elimination and enhancement of user experience aiming for games that connect with them on a deeper level. Parents can control the influence of certain games on their kids and track the emotions associated with them.

The study has certain limitations. Adult's face differs from a child's, but we made training on all kinds of faces. Moreover, interpreting emotions solely through audio and video signals might not capture the complete emotional context of gameplay, because children of different ages might have varying emotional responses and cognitive abilities. Despite its limitations, preliminary results demonstrate that it is a promising tool that has the potential to make computer games more child-oriented based on emotional data.

The study has several open questions. In particular, we want to understand more about how emotions from sound and video connect and contrast with each other. Researchers like \cite{fusion1}, \cite{fusion2}, \cite{fusion3}, \cite{attention}, \cite{icanet} have delved into similar areas.

As for future work, we plan to test the system in real settings to see how well it performs. Future experiments will involve more participants of different ages and engaging with different game types. After testing, it will be possible to conduct interviews with the participants to ask clarifying questions. According to recent findings, the expression of fear and neutral emotions between adults and children is quite different between kids and adults \cite{app12167992}. So, we plan to improve the ER framework by training the models on kids' faces and sounds only.





\bibliographystyle{elsarticle-harv}
\bibliography{export}

\clearpage


\end{document}